\documentclass[a4paper,11pt]{article}
\pdfoutput=1
\usepackage{jheppub}
\usepackage{rotating}
\usepackage{array}
\usepackage{amsmath}
\usepackage[normalem]{ulem}
\usepackage{slashed}
\usepackage{booktabs}
\usepackage[pdftex,table]{xcolor}
\usepackage{units}
\usepackage{xfrac}
\usepackage{mathtools}
\usepackage{empheq}
\usepackage[]{units}
\usepackage{multirow}
\usepackage{amssymb}
\usepackage{url}
\usepackage{comment}
\usepackage{paralist}

\usepackage{tikz}

\def\beq{\begin{equation}}
\def\eeq{\end{equation}}
\newcommand{\bea}{\begin{eqnarray}\begin{aligned}}
\newcommand{\eea}{\end{aligned}\end{eqnarray}}
\def\bitem{\begin{itemize}}
\def\eitem{\end{itemize}}

\definecolor{darkpurple}{rgb}{0.5, 0.2, 0.8}
\definecolor{darkblue}{rgb}{0.0, 0.0, 0.8}
\definecolor{darkgreen}{rgb}{0.0, 0.4, 0.0}
\definecolor{darkred}{rgb}{0.5, 0.0, 0.0}
\usepackage{hyperref}
\hypersetup{
    linktocpage,
     colorlinks,
     citecolor=darkgreen,
     linkcolor= darkgreen,
     urlcolor=darkgreen
}
\abstract{
Many artificial intelligence (AI) devices have been developed to accelerate the training and inference of neural network models. The most common ones are the Graphics Processing Unit (GPU) and Tensor Processing Unit (TPU). They are highly optimized for dense data representations. However, sparse representations such as graphs are prevalent in many domains, including science. It is therefore important to characterize the performance of available AI accelerators on sparse data. This work analyzes and compares the GPU and TPU performance training a Graph Neural Network (GNN) developed to solve a real-life pattern recognition problem. Characterizing the new class of models acting on sparse data may prove helpful in optimizing the design of deep learning libraries and future AI accelerators. 
}


\begin{document}

\title{\boldmath Benchmarking GPU and TPU Performance with Graph Neural Networks}

\author[1]{Xiangyang Ju,}
\author[2]{Yunsong Wang,}
\author[3]{Daniel Murnane,}
\author[3]{Nicholas Choma,}
\author[2]{Steven Farrell} 
\author[3]{and Paolo Calafiura}
\affiliation[1]{Physics Division, Lawrence Berkeley National Laboratory, Berkeley, CA 94720, USA}
\affiliation[2]{NERSC, Lawrence Berkeley National Laboratory, Berkeley, CA 94720, USA}
\affiliation[3]{Scientific Data Division, Lawrence Berkeley National Laboratory, Berkeley, CA 94720, USA}
\emailAdd{xju@lbl.gov}
\emailAdd{yunsongwang@lbl.gov}
\emailAdd{dtmurnane@lbl.gov}
\emailAdd{njchoma@lbl.gov}
\emailAdd{sfarrell@lbl.gov}
\emailAdd{pcalafiura@lbl.gov}

\maketitle
\flushbottom

\section{Introduction} \label{sec:intro}
Modern machine learning (ML) plays a critical role in numerous domains, including computer vision, language processing, and speech/image recognition. Much of their success is driven by three factors: novel deep learning models, large-scale datasets, and massive computing power. ML models are getting wider and deeper, and reaching a trillion trainable parameters~\cite{brown2020language}. To keep up with the demands of deep learning at the end of Moore’s Law, novel specialized computing devices, collectively known as AI accelerators, are necessary. 

The most popular AI accelerator is the Graphics Processing Unit (GPU). GPUs are optimized for massively parallel execution of simple code blocks, and well-suited for linear algebra. Similarly, the Tensor Processing Unit (TPU) is optimized for matrix operations~\cite{tpu-eval}. Both GPU and TPU are optimized to operate on dense matrices.

MLPerf~\cite{mlperf} is a machine learning benchmark suite that has gained industry-wide support and recognition. It includes computer vision, language processing, recommendation system and gaming applications. There are other ML benchmark being proposed such as the  ParaDNN~\cite{wang2019benchmarking}.
which focuses on fully connected, convolutional and recurrent neural networks. 
These benchmarks, while essential for fair comparisons among different architectures, do not capture the performance characteristic of models (such as Graph Neural Networks) that operate on sparse and irregular datasets. Sparse datasets are common in science applications such as molecular dynamics, genomics, and High Energy Physics (HEP). Graph representation and GNNs have seen rapidly growing applications in these science domains. Ref~\cite{Shlomi:2020gdn} reviews GNN application to HEP. This work uses as a benchmark a GNN model that solves a combinatorially hard pattern recognition problem on the Large Hadron Collider data. 


This paper is organized as follows. Section~\ref{sec:gnn} describes the benchmark GNN model and the dataset used for training it. Section~\ref{sec:devices} introduces the hardware platforms used for this study. 
Section~\ref{sec:comparison} provides a thorough comparison of TPU and GPU. The performance analysis is described in Section~\ref{sec:perf}. Outlook and conclusions are in Section~\ref{sec:conclusions}.

\section{Graph Neural Network Benchmark} \label{sec:gnn}
\begin{figure}[htb]
    \centering
    \includegraphics[width=0.65\textwidth]{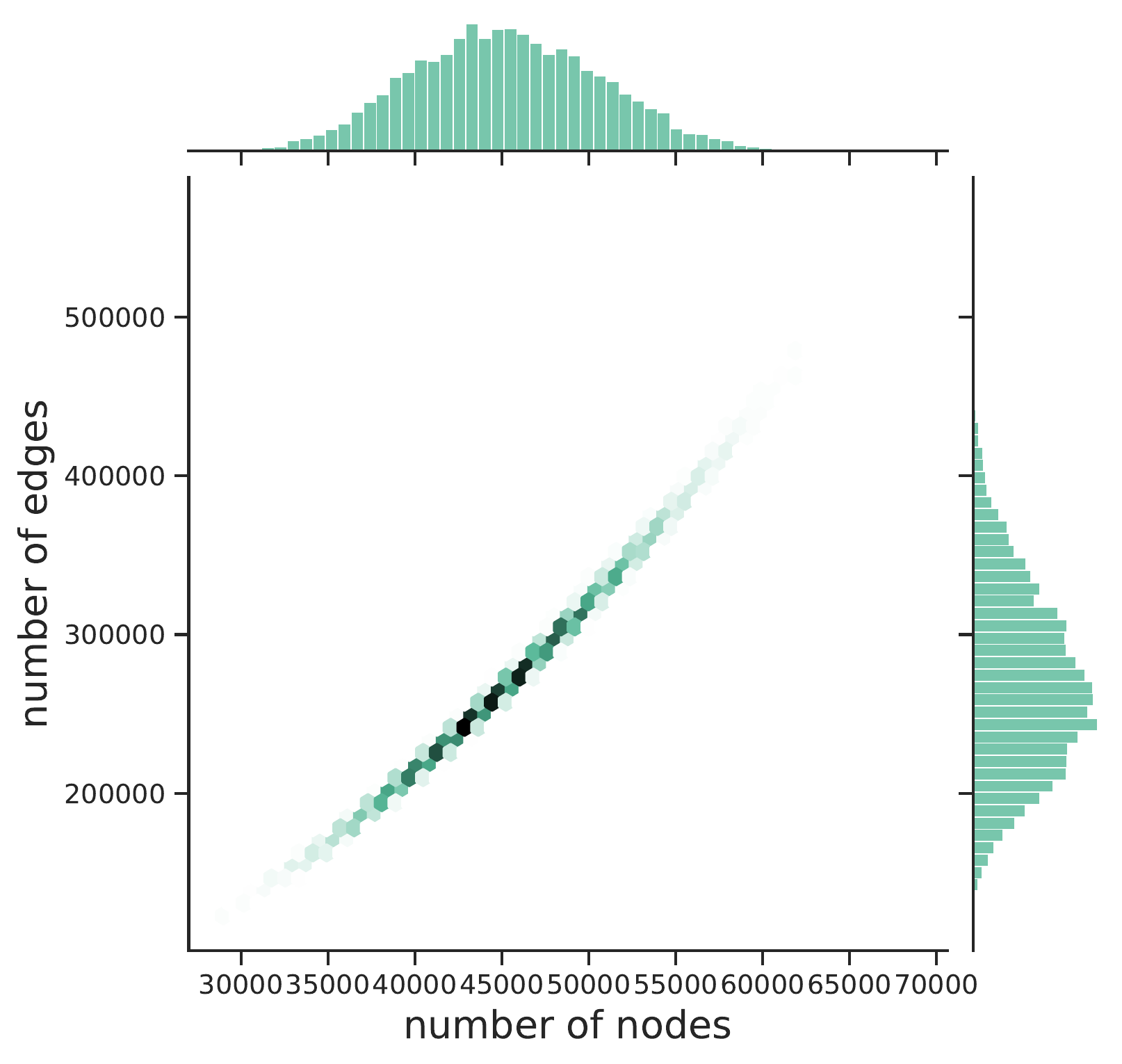}
    \caption{Number of nodes and edges in the input graphs.}
    \label{fig:graph_size}
\end{figure}

The data are from the TrackML~\cite{trkML,Amrouche:2019wmx} dataset, which simulates top quark pair production from proton-proton collisions at the Large Hadron Collider. Graphs are constructed with the embedding learning and filtering method described in Ref~\cite{choma2020track}. 
In this application, the graph contains all data from a collision event. Each node represents a 3D space-point measurement ({\it hit}), and each edge represents a directional connection of one hit (sender) pointing to another (receiver). Hit positions in the cylindrical coordinate system are assigned as node attributes. The objective of graph neural networks is to assign a score to each edge so as to indicate the probability that the edge is {\it true}, i.e. that it connects two hits from the same particle. The graph size varies for each collision event, as shown in Figure~\ref{fig:graph_size}. On average, there are about 50,000 nodes and 250,000 edges. Out of the 250,000 edges, the expected number of true edges is about 50,000.

The GNN architecture has three trainable components~\cite{Ju:2020xty}: an encoder that transforms input graphs to their latent representations, an interaction network~\cite{battaglia2018relational} and a decoder that computes the edge scores. The encoder has two independent networks: node network and edge network, which use two layers of fully connected network with sizes of [128, 64]. The interaction network also has node and edge networks. It first updates node features and then updates the edge features. Node features are updated with a multilayer perceptron (MLP) to which the inputs are the concatenated node features and the aggregated neighboring edge features. Edge features are updated with another MLP to which the inputs are the concatenated edge features and the sender/receiver node features. In this way, messages are passed between nodes and edges via a message passing operation, one of which is the \textsc{UnsortedSegmentSum} in TensorFlow~\cite{abadi2016tensorflow}. After performing the update eight times, edge features are fed to the decoder to predict edge scores. There are 132,291 trainable parameters in the GNN. 


\section{Hardware and Software} \label{sec:devices}
Our selection of hardware reflects the latest configurations available on the Google Cloud Platform (GCP) at the time of writing. Table~\ref{tab:hardwares} lists basic  parameters for each AI accelerator. The hourly cost to use each device is the GCP list price.

\begin{table}[htb]
    \centering
    \begin{tabular}{c|c|c|c|c|c|c}
    \hline
         Device &  Architecture & chips & Peak Flops & High-Bandwidth  & Cost & Thermal Design \\
          & & & [TFLOPS] & Memory [GiB] & [USD/hour] & Power [W]\\
          \hline
         GPU & Volta & 1 & 14 (fp32) & 16 & 1.56 & 250 \\
         GPU & Ampere & 1 & 19.5 (fp32) & 40 & N/A & 250 \\
         TPU & v2 & 32 & 720 & 256 & 15.33 & 2400 \\
         TPU & v3 & 8 & 420 & 128 & 8 & 600 \\
         \hline
    \end{tabular}
    \caption{Technical parameters for AI accelerators accessible on the Google Cloud Platform. Those TPU metrics are the summed metrics of all chips in the TPU.}
    \label{tab:hardwares}
\end{table}

The GPU is an NVIDIA V100 in a computing node at the National Energy Research Scientific Computing Center (NERSC). Each computing node contains eight V100 packages (PCIe) connected via 25 GB/s NVlink connection, and each V100 armed with 5120 CUDA cores has 16 GB of memory and 900 GB/s memory bandwidth. It reaches peak performance of 14 TFLOPS in single precision (float32) and seven teraflops in double precision (float64). In addition, A V100 has 640 tensor cores and can run mixed-precision training using half-precision (float16) to compute and single precision to accumulate, making its peak performance 125 TFLOPS. However, studies are needed to make sure the model converges to optimal results. Tensor cores are not used in this study. In May 2020, Nvidia announced the Nvidia A100, to which we were granted early beta access on GCP. The peak performance in single precision of the Nvidia GPU A100 is about 40\% better than that of Nvidia GPU V100. Our experiments comparing the GNN training time between V100 and A100 are consistent with the expected peak performance improvement. Only the results with V100 are presented here. 

The TPU is a Cloud TPU instance to which we were given academic access in Spring 2020. TPUs are designed to run whole inference models in the TPU to reduce communications with the host CPU. Details of TPU architectures can be found in Ref~\cite{tpu-eval}. The heart of the TPU is the Matrix Unit (MXU). It can perform 16,000 multiply-accumulate operations in each cycle at reduced precision (bfloat16). It supports mixed-precision training, using bfloat16 to compute and float32 to accumulate. There are two versions of TPU cores: v2 and v3. A TPU v2 core has 8 GB of memory and one MXU, while a TPU v3 core doubles the memory size and MXU. However, the number of cores in each TPU can be configured~\cite{tpu-types}. We chose 32 v2 cores and eight v3 cores in the GCP {\tt us-central1-a} region for this study.

The software stack is based on the TensorFlow v2.3~\cite{abadi2016tensorflow}, which unifies the usage of CPUs, GPUs, and TPUs via the \textsc{tf.distribute} package. It allows users to perform data distributed training for all kinds of devices in a consistent way. In addition, we use \textsc{tf.data} to build TensorFlow input pipelines, and use \textsc{tf.function} to compile the model and generate a computation graph in gRPC format~\cite{grpc}. The gRPC file is sent via the cloud to the TPU host and compiled by XLA, and in the end, the binary files are executed in TPUs. As of writing~\footnote{In 10/2020.}, the TPU does not support dynamic graph sizes. The input graph size has to be the same. Padding all graphs to the largest graph in the training data is expensive. Instead, we padded to the graph whose number of nodes and edges have a quantile of 99\%, saving the training time by 30\% compared to that padding to the largest graph size. 

\section{Comparison of TPU and GPU} \label{sec:comparison}
In this section, we use the following key metrics to compare the computational performances of TPUs and GPUs: accuracy, latency, cost, power consumption. 

\begin{figure}[htb]
\begin{minipage}[b]{.49\textwidth}
  \centering
  \includegraphics[width=\textwidth]{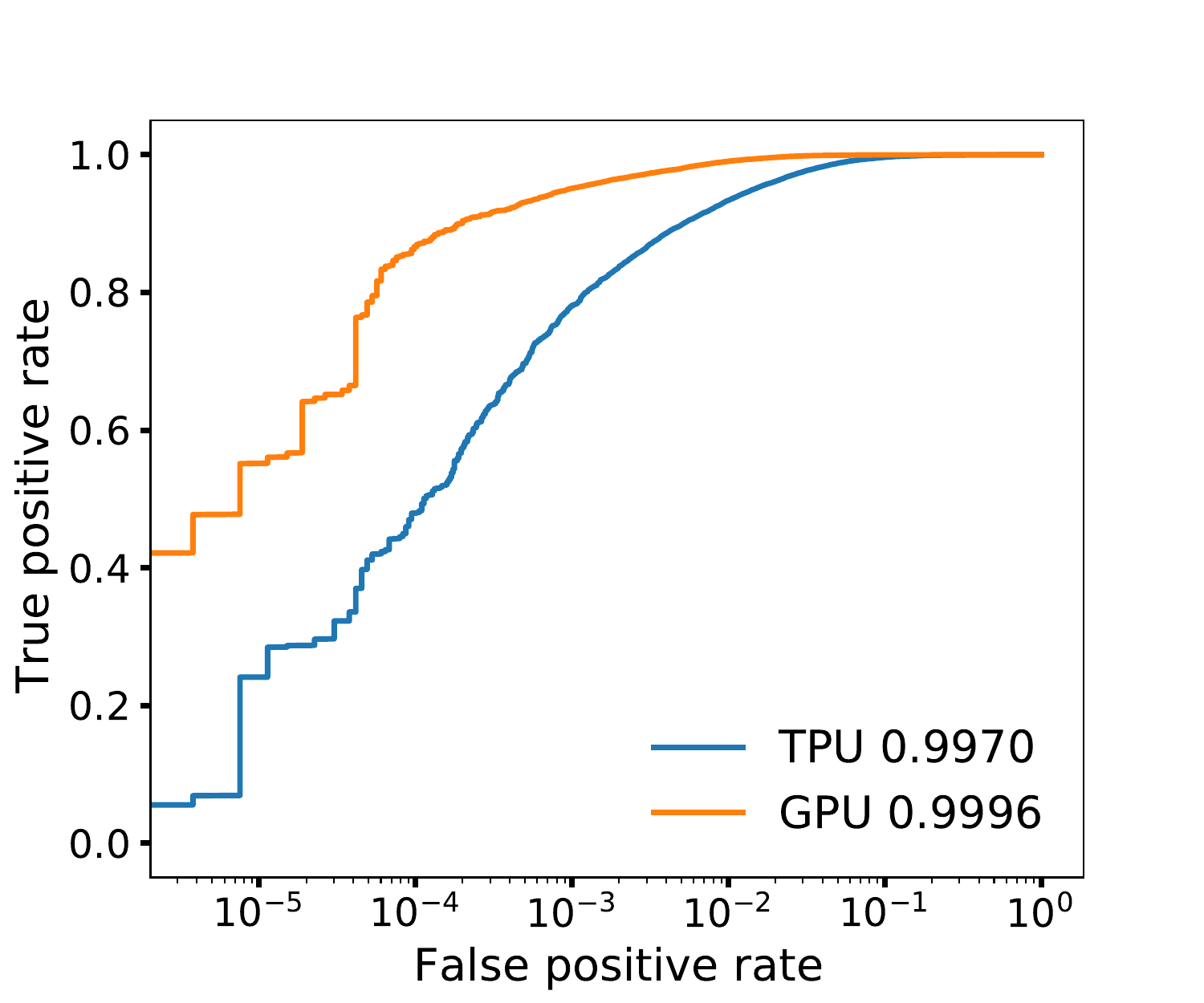}
  \caption{The AUC for TPU and GPU.}
  \label{fig:auc}
\end{minipage}%
\hfill
\begin{minipage}[b]{.49\textwidth}
  \centering
  \includegraphics[width=\textwidth]{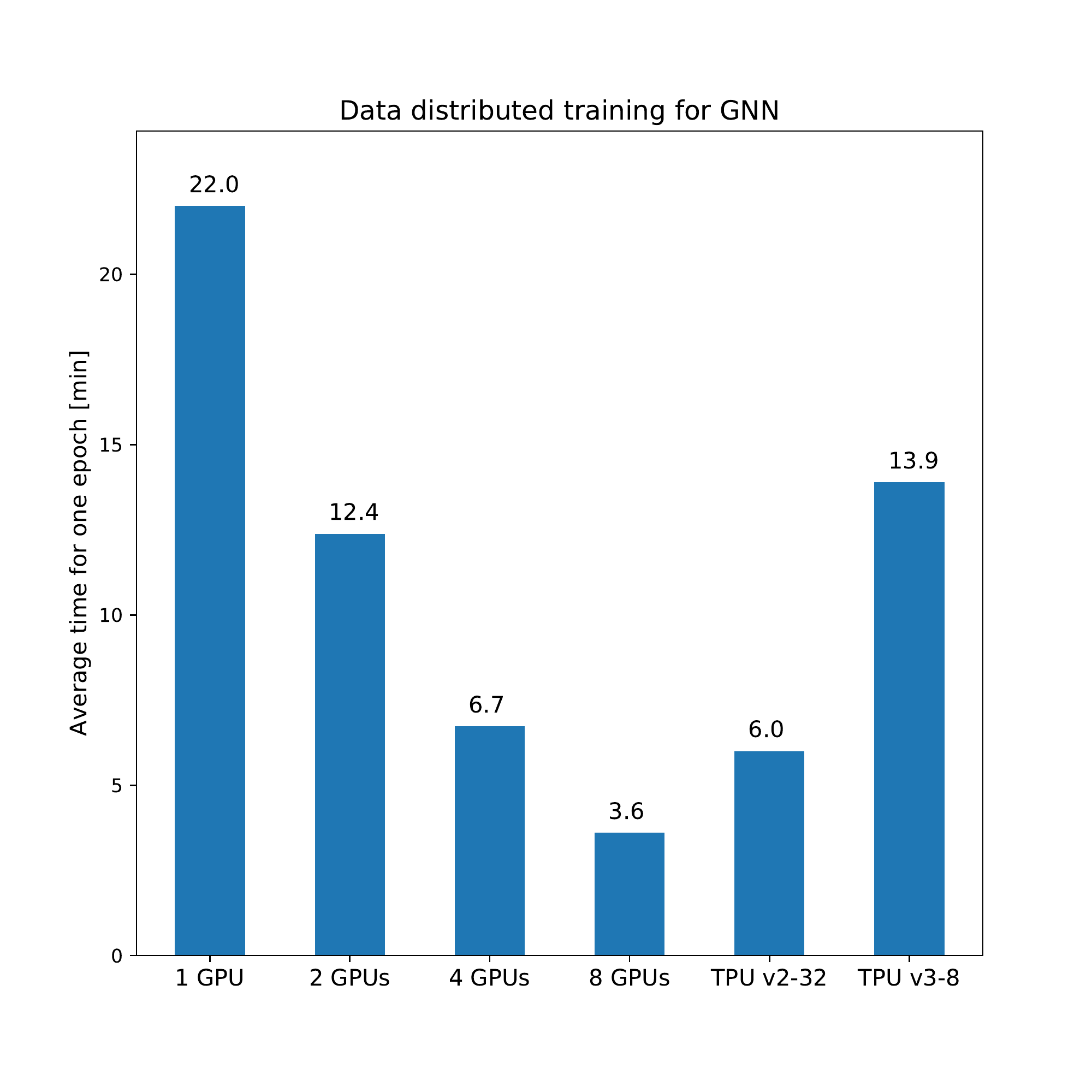}
  \caption{The averaged time for one training epoch for TPU and GPU.}
  \label{fig:latency}
\end{minipage}
\end{figure}

\paragraph{Accuracy}  It measures the quality of the trained model and is evaluated by the Area Under the receiver operating characteristic curve (AUC). Figure~\ref{fig:auc} shows the comparison of accuracy achieved with TPU and GPU. The difference between the two training results lies in the batch size. Training with GPU uses a batch size of 1, while training with TPU v3 uses 8. GPU and TPU performances are very close, even though model hyperparameters, particularly the learning rate, were not tuned for TPUs.

\paragraph{Latency} It measures the time it takes to finish one training epoch. It is one of the critical metrics for HEP online data processing. 
Figure~\ref{fig:latency} shows the comparison of the averaged time it takes to finish one training epoch. A TPU with 32 TPU v2 cores runs as fast as 4 GPU V100 does and A TPU with 8 TPU v3 cores runs as fast as 2 GPU V100 does. 

\begin{figure}[htb]
\begin{minipage}[b]{.49\textwidth}
  \centering
  \includegraphics[width=\textwidth]{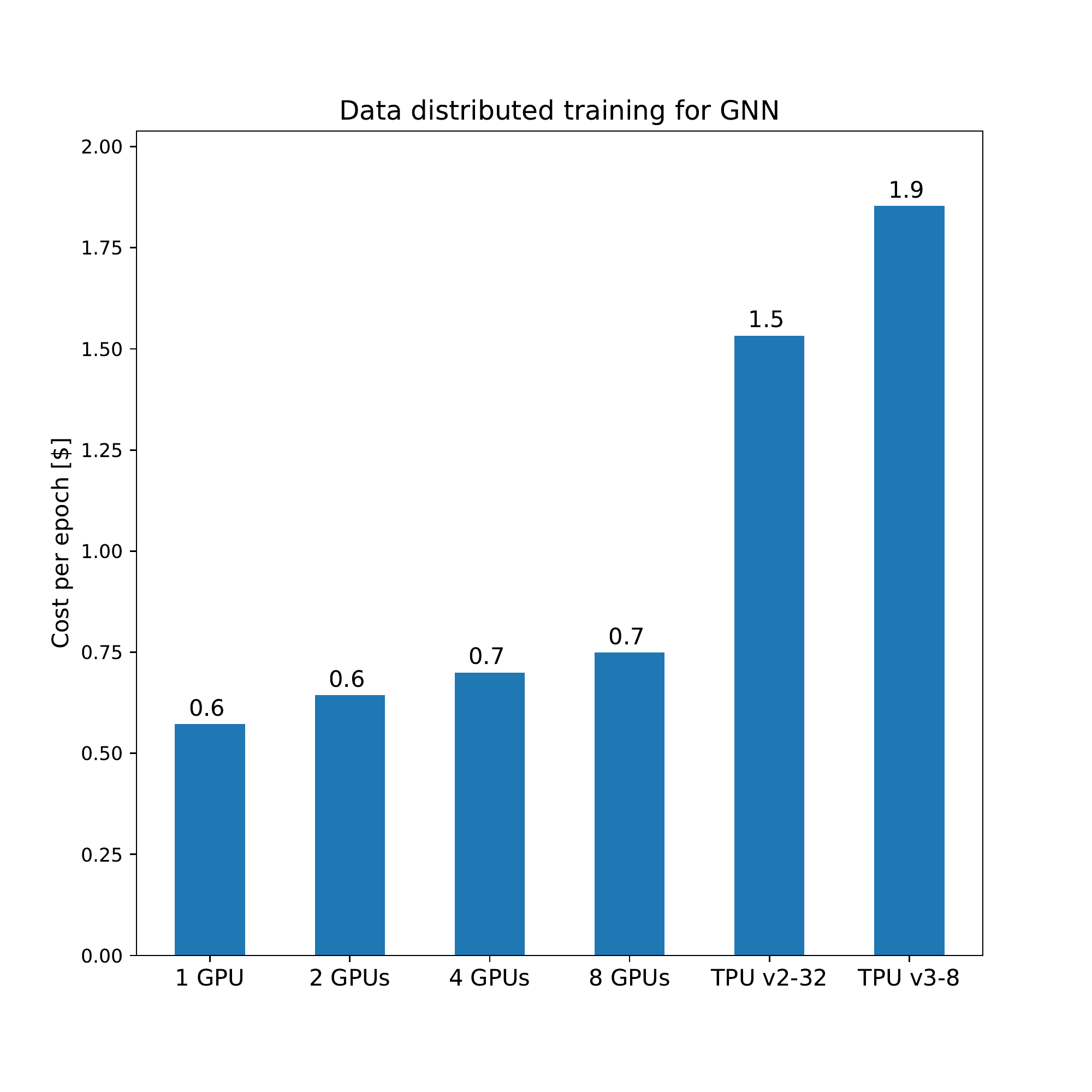}
  \caption{The cost for one training epoch for TPU and GPU.}
  \label{fig:cost}
\end{minipage}%
\hfill
\begin{minipage}[b]{.49\textwidth}
  \centering
  \includegraphics[width=\textwidth]{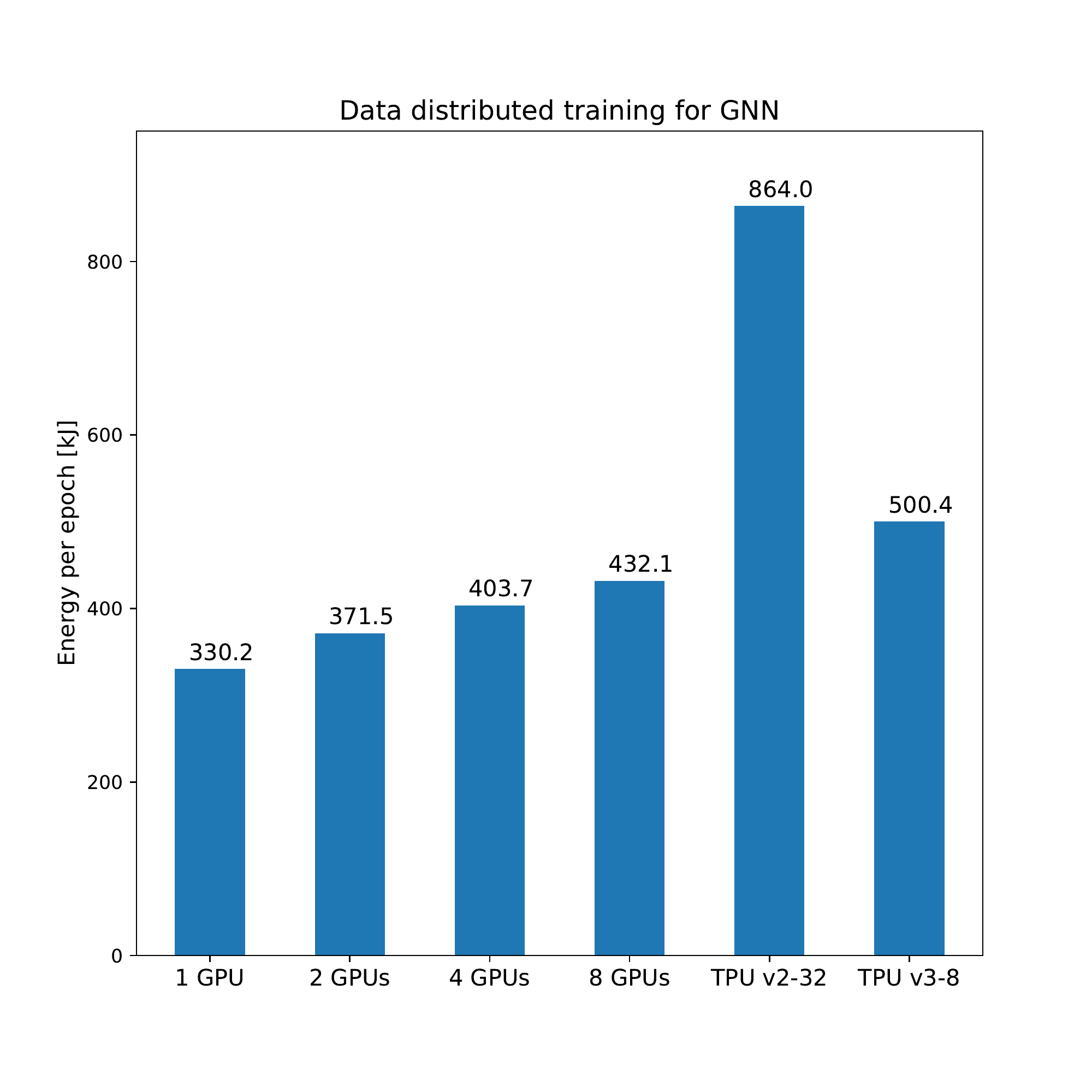}
  \caption{The power consumption for one training epoch for TPU and GPU.}
  \label{fig:power}
\end{minipage}
\end{figure}

\paragraph{Cost} It measures the money it costs to train one epoch in US dollars, as shown in Figure~\ref{fig:cost}. The prices for using the devices are taken from Google Cloud Platform, charged by US dollars per hour. It is found that using GPUs is more economical than using TPUs. Simultaneously using multiple GPUs for training will increase the cost a bit due to the imperfect strong scaling efficiency.

\paragraph{Power Consumption} It measures the energy cost per epoch, a key metric to evaluate if the device is environment friendly. It is calculated as the thermal design watts times the latency, as shown in Figure~\ref{fig:power}. For simplicity, we assume the device is 100\% busy during the training, which usually does not hold. It is found the GPU is more environmentally friendly.

\section{Profiling and Roofline analysis } \label{sec:perf}
To understand the results shown in Section~\ref{sec:comparison}, we use Tensor Board v2.3~\cite{abadi2016tensorflow} to profile the training process for TPU v3 and GPU. This section describes the profiled results and the performance analysis with the Roofline model~\cite{roofiline}.

We break the time usage of each device into different TensorFlow operations. For TPU v3, 37.7\% of total training time is spent in \textsc{UnsortedSegmentSum}, which aggregates the node or edge features,  13.6\% in idling, and  12.7\% in matrix multiplications. On the other hand, for GPU,  37.4\% of total training time is spent in matrix multiplications,  10\% in idling, and 6.4\% in \textsc{UnsortedSegmentSum}. In both cases, about half of the time is spent in gradient calculations and back-propagation.

To determine which kernel takes most of the training time, we ranked all kernels in descending order of their total duration in each training step. The top three CUDA kernels account for 98.7\% of total FLOPs but only for 40.7\% of total training time. They are to perform matrix multiplications. The following two CUDA kernels account for 13.8\% of total training time but zero FLOPs. Specifically, the two CUDA kernels are to concatenate and slice tensors. The top 5 to 20 ranked kernels are led by the message-passing operation \textsc{UnsortedSegmentSum}.

FLOP utilization, defined as the fraction of the FLOPS used over the peak FLOPS offered by the device, is an essential metric in understanding how well an application utilizes the accelerator’s total computation capacity. The GNN trained with TPU v3 has a FLOP utilization of 2.3\%, but that with GPU V100 has a FLOP utilization of about 30\% for single precision. The TensorCore in V100 was not used because it requires the benchmark to support mixed-precision training. 

\begin{figure}
    \centering
    \includegraphics[width=0.49\textwidth]{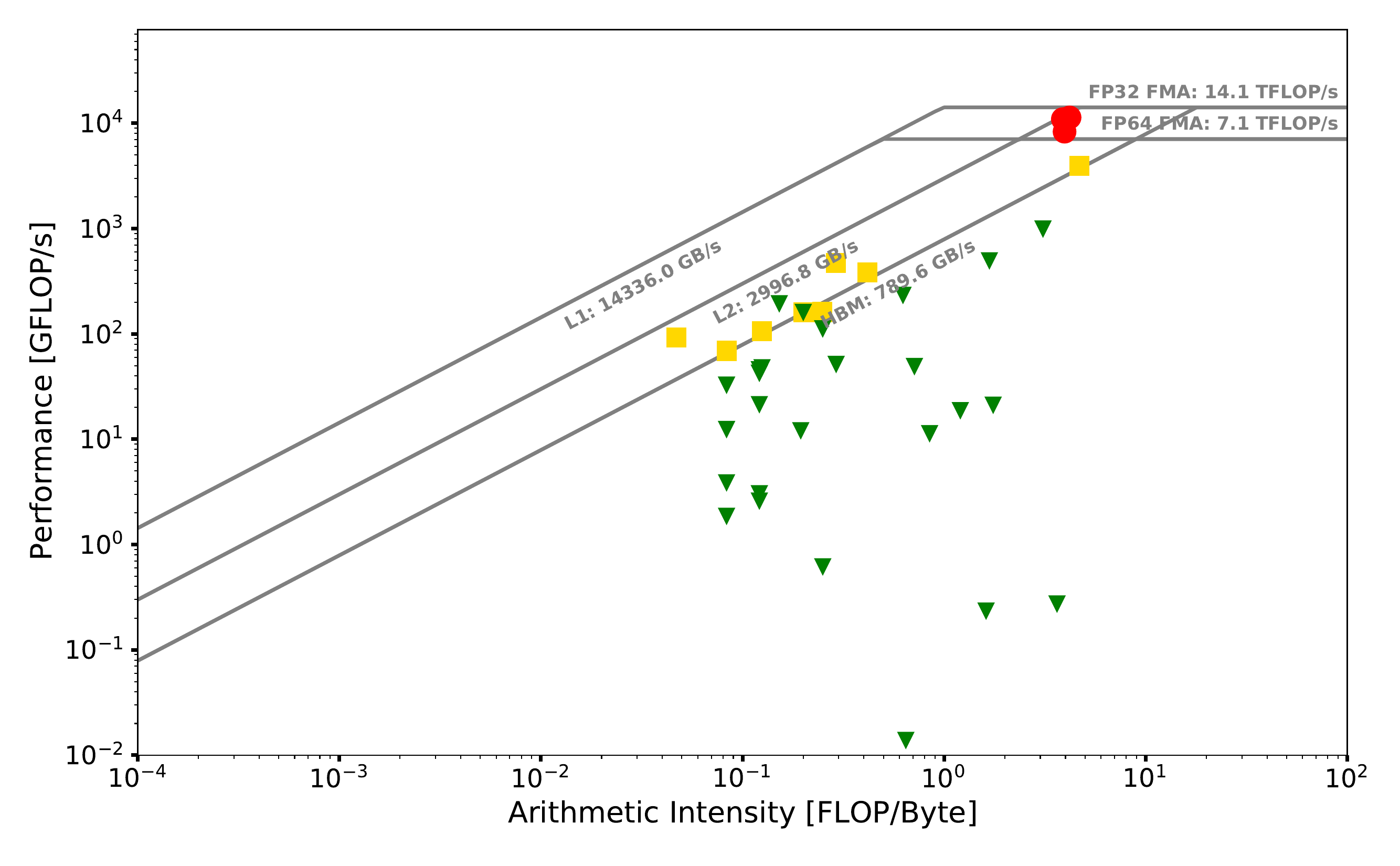}
    \includegraphics[width=0.49\textwidth]{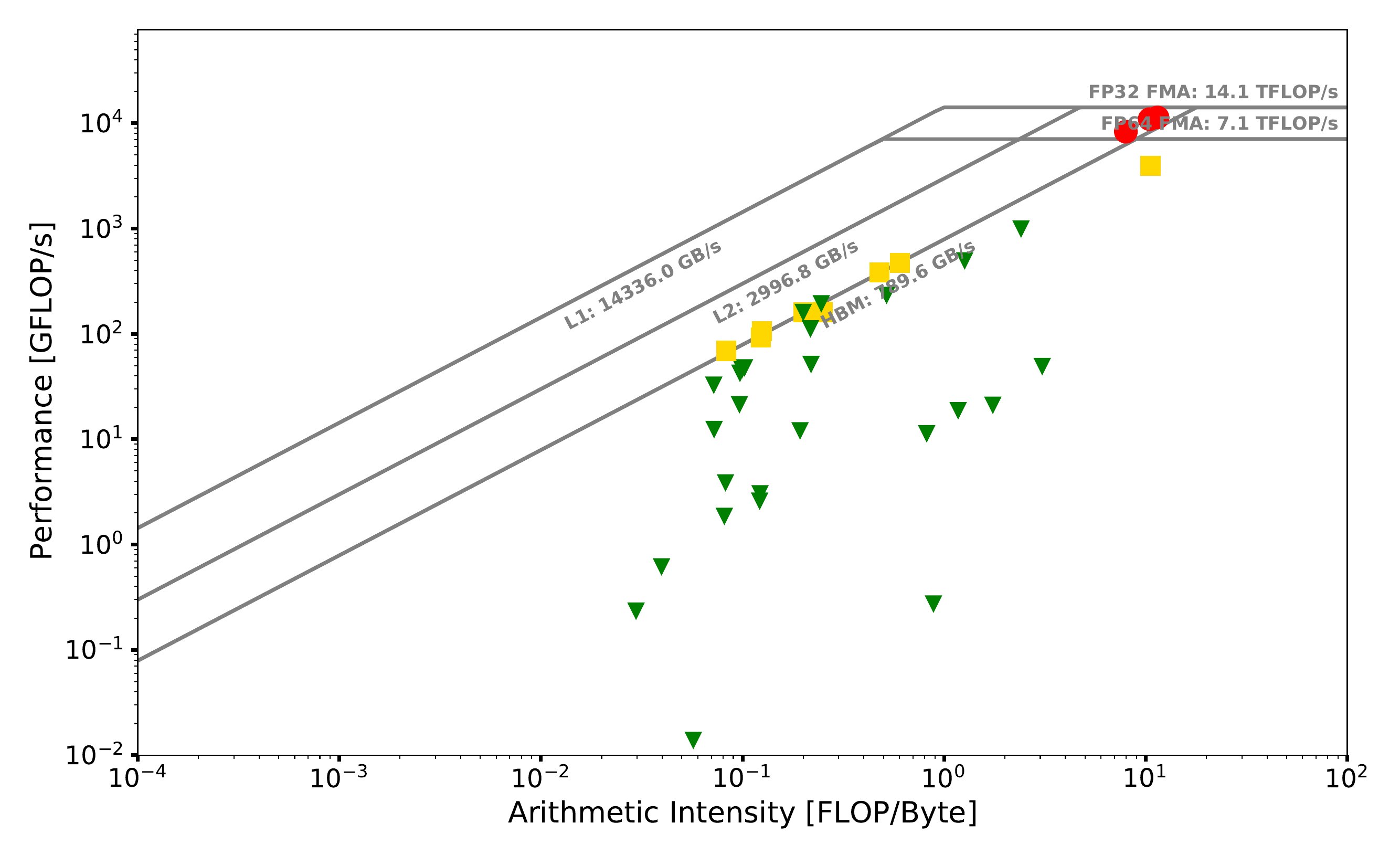}
    \includegraphics[width=0.49\textwidth]{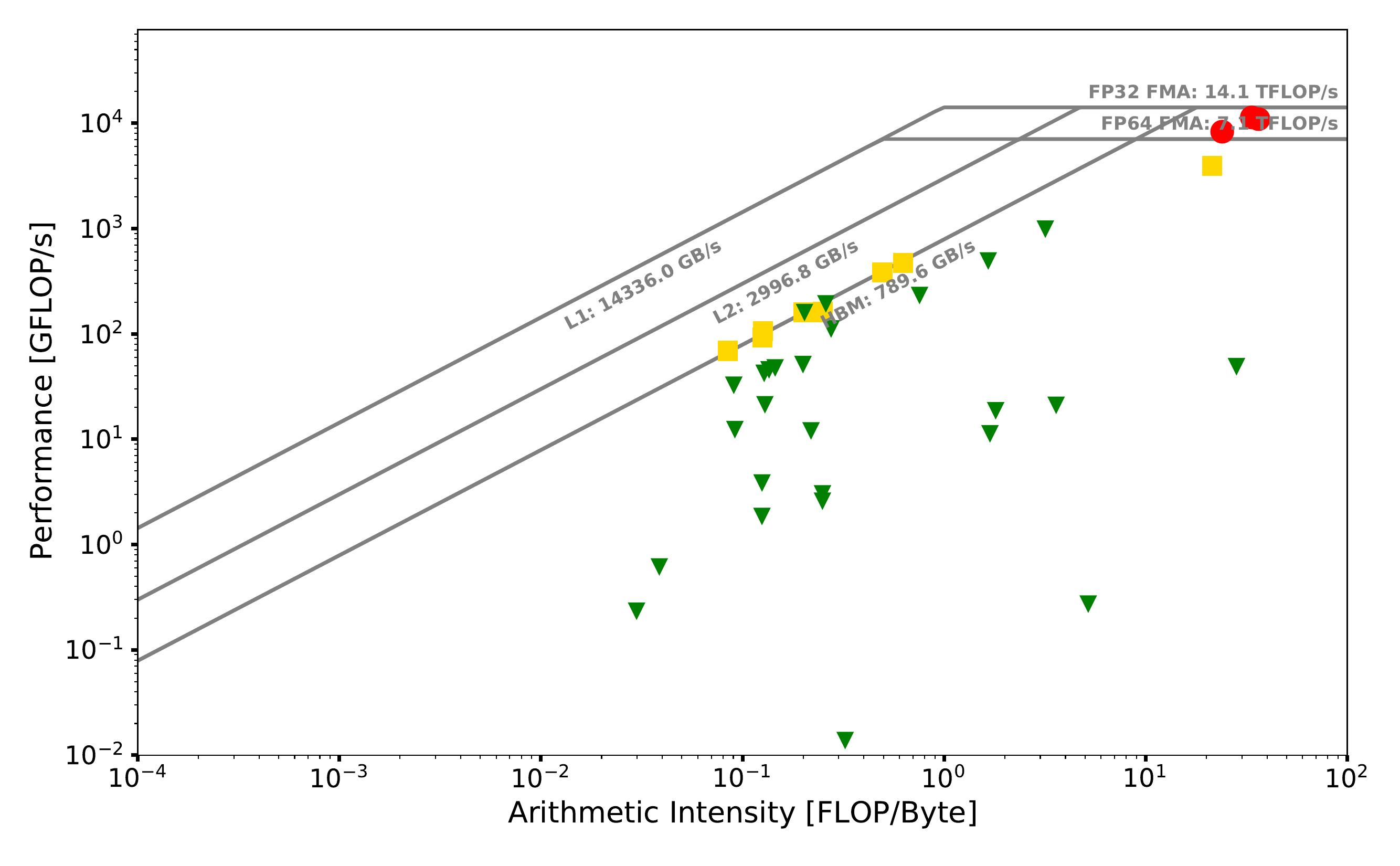}
    \includegraphics[width=0.49\textwidth]{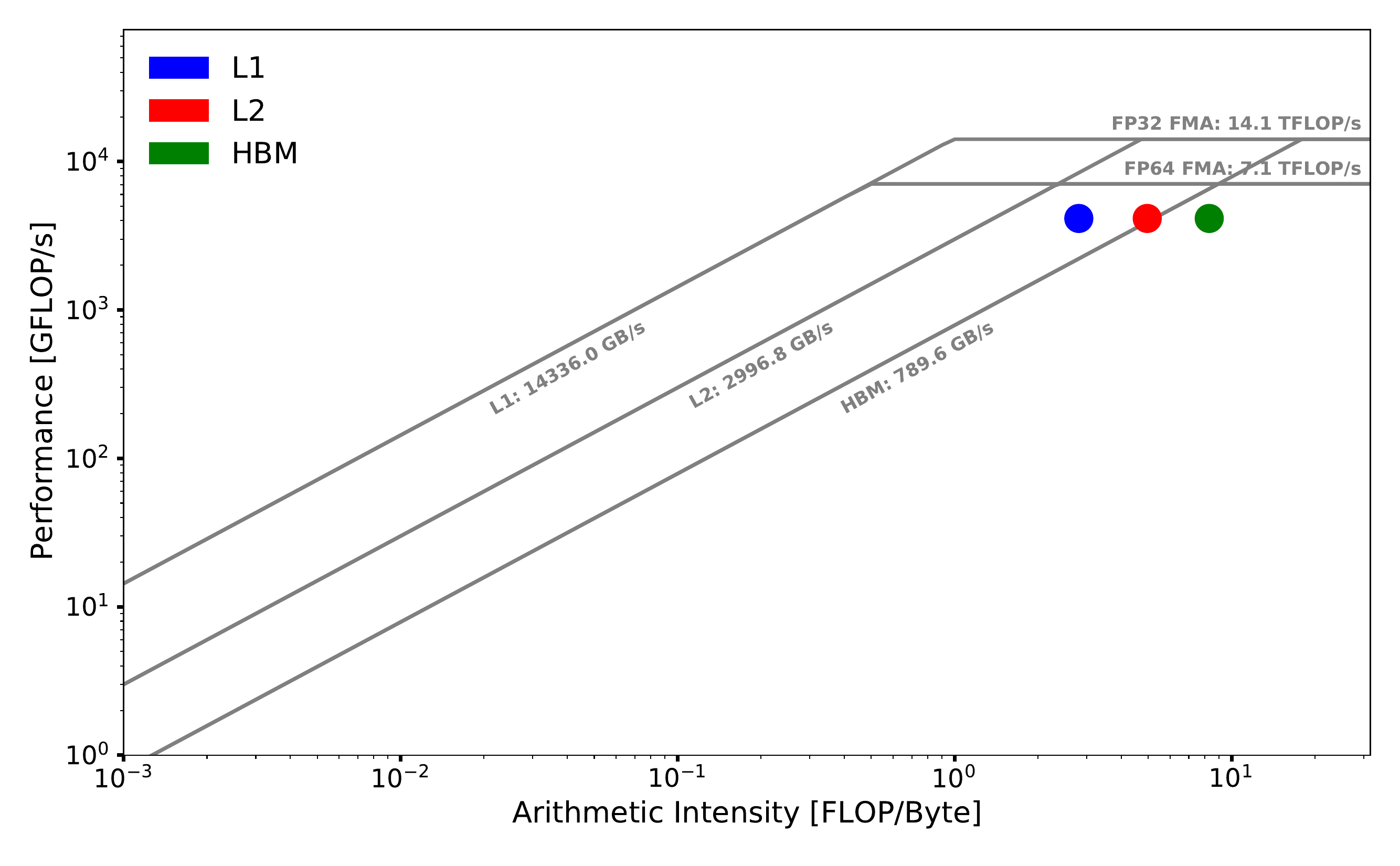}
    \caption{Performance analysis with the roofline model for different memory hierarchical level: L1 memory (top left), L2 memory (top right), High bandwidth memory (bottom left) and a summary (bottom right). In all plots except the summary one, the red dots represent the top 5 CUDA kernels, the yellow squares the top (5 - 20] CUDA kernels and the green triangle the top 20+ CUDA kernels. These kernels are ranked by the time each kernel runs in each training step.}
    \label{fig:roofline}
\end{figure}

FLOPS utilization is only part of the problem when designing an accelerator. In particular, memory bandwidth is another important aspect that can have a significant impact on performance. We use the Roofline model to study the GNN training runs for GPU V100. Figure~\ref{fig:roofline} shows the Roofline studies for different memory hierarchical levels. The Roofline has a slanted part and a horizontal part. It represents the highest achievable FLOPS at a given arithmetic intensity. Any data point ($x$, $y$) on the slanted part has $y/x$ = memory bandwidth (GB/s). The horizontal part is the peak FLOPS on the device. A workload, which is a CUDA kernel in our case, close to the slanted roofline is memory-bound; one close to the horizontal part is compute-bound. Figure~\ref{fig:roofline} (bottom left) shows that the kernels performing message-passing operations indicated in yellow squares are bounded by the high bandwidth memory. It is the interest of future device design to improve memory bandwidth and add sparse operation support. The Roofline model ignores the CUDA kernels with zero arithmetic intensity (zero-AI kernels). The zero-AI kernels, summing to 1/3 of total CUDA kernels, take 44.8\% of total training time. They account for 20.6\% of data transactions in L1, 38.0\% in L2, and 54.0\% in HBM. The total overhead in creating all zero-AI kernels is about 2.4\% of the total training time.

\clearpage
\section{Conclusions} \label{sec:conclusions}
Graph Neural Networks combine deep learning operations (e.g., matrix multiplication and convolution) with iterative message passing operations such as \textsc{UnsortedSegementSum}, which aggregates node (edge) features to edge (node) networks. GNN is a unique group of underrepresented neural networks in current popular benchmark suites such as MLPerf.

We use a GNN model that solves a scientific problem as a benchmark to compare the computational performances of GPU and TPU. In order to utilize more than one TPU core, the distributed training of GNN was implemented. Different software implementations result in different training time for one epoch and different strong scaling efficiencies. The GNN model’s accuracy trained with TPU and GPU is very similar; however, they manifest different latencies. A TPU with 32 TPU v2 cores runs as fast as 4 GPU V100 does. TPUs spend most of the training time in running message passing operations but GPUs in running matrix multiplications. GPUs cost less money and consume less power compared with TPUs. The Roofline analysis reveals that the bottlenecks of training GNN on GPUs are the computing capability for the matrix operations and the memory bandwidth for the message passing operations. Besides GPUs and TPUs, Field Programmable Gate Arrays (FPGA)~\cite{intelFPGA} and Intelligence Processing Units (IPU)~\cite{jia2019dissecting} offer alternative approaches to AI acceleration that show promise for sparse data. It would be interesting for future studies to include IPUs and FPGAs in the comparison.

\acknowledgments
We would like to thank Zachary Marshall and Wahid Bhimji for their useful feedback on the manuscript.

This research used resources of the National Energy Research Scientific Computing Center (NERSC), a US Department of Energy Office of Science User Facility operated under Contract No. DE-AC02-05CH11231.

We are grateful to the Google Cloud Platform team, and in particular to Ema Kaminskaya for the access to TPU and Nvidia A100 resources in the context of the GCP/US ATLAS collaboration. 

\bibliography{graph}

\providecommand{\href}[2]{#2}\begingroup\raggedright\begin{thebibliography}{10}

\bibitem{brown2020language}
T.~B. Brown, B.~Mann, N.~Ryder, M.~Subbiah, et~al., {\it Language models are
  few-shot learners},  2020.

\bibitem{tpu-eval}
N.~P. Jouppi et~al., {\it In-datacenter performance analysis of a tensor
  processing unit},  {\em SIGARCH Comput. Archit. News} {\bf 45} (June, 2017)
  1–12, [\href{http://arxiv.org/abs/1704.04760}{{\tt arXiv:1704.04760}}].

\bibitem{mlperf}
MLPerf, ``Fair and useful benchmarks for measuring training and inference
  performance of ml hardware, software and services.''
  \url{https://mlperf.org/}.

\bibitem{wang2019benchmarking}
Y.~E. Wang, G.-Y. Wei, and D.~Brooks, {\it Benchmarking tpu, gpu, and cpu
  platforms for deep learning},  2019.

\bibitem{Shlomi:2020gdn}
J.~Shlomi, P.~Battaglia, and J.-R. Vlimant, {\it {Graph Neural Networks in
  Particle Physics}},  \href{http://arxiv.org/abs/2007.13681}{{\tt
  arXiv:2007.13681}}.

\bibitem{trkML}
``{TrackML Particle Tracking Challenge}.''
  \url{https://www.kaggle.com/c/trackml-particle-identification}.

\bibitem{Amrouche:2019wmx}
S.~Amrouche et~al., {\it {The Tracking Machine Learning challenge : Accuracy
  phase}},  \href{http://arxiv.org/abs/1904.06778}{{\tt arXiv:1904.06778}}.

\bibitem{choma2020track}
N.~Choma, D.~Murnane, X.~Ju, P.~Calafiura, et~al., {\it {Track Seeding and
  Labelling with Embedded-space Graph Neural Networks}},
  \href{http://arxiv.org/abs/2007.00149}{{\tt arXiv:2007.00149}}.

\bibitem{Ju:2020xty}
X.~Ju et~al., {\it {Graph Neural Networks for Particle Reconstruction in High
  Energy Physics detectors}},  in {\em {33rd Annual Conference on Neural
  Information Processing Systems}}, 3, 2020.
\newblock \href{http://arxiv.org/abs/2003.11603}{{\tt arXiv:2003.11603}}.

\bibitem{battaglia2018relational}
P.~W. Battaglia, J.~B. Hamrick, V.~Bapst, A.~Sanchez-Gonzalez, et~al., {\it
  {Relational inductive biases, deep learning, and graph networks}},
  \href{http://arxiv.org/abs/1806.01261}{{\tt arXiv:1806.01261}}.

\bibitem{abadi2016tensorflow}
M.~Abadi et~al., {\it Tensorflow: Large-scale machine learning on heterogeneous
  distributed systems},  \href{http://arxiv.org/abs/1603.04467}{{\tt
  arXiv:1603.04467}}.

\bibitem{tpu-types}
Google, ``Tpu types and zones.''
  \url{https://cloud.google.com/tpu/docs/types-zones}.

\bibitem{grpc}
gRPC Authors, ``grpc.'' \url{https://grpc.io/}.

\bibitem{roofiline}
S.~Williams, A.~Waterman, and D.~Patterson, {\it Roofline: an insightful visual
  performance model for multicore architectures},  vol.~52, pp.~65--76,
  Communications of the ACM, 4, 2009.

\bibitem{intelFPGA}
B.~G. Richard~Chuang, Omi~Oliyide, ``Introducing the intel vision acceleator
  design with intel arria 10 fpga.''
  \url{https://www.intel.com/content/dam/www/programmable/us/en/pdfs/literature/wp/intel-vision-accelerator-design-with-FPGA-wp.pdf},
  2019.

\bibitem{jia2019dissecting}
Z.~Jia, B.~Tillman, M.~Maggioni, and D.~P. Scarpazza, {\it Dissecting the
  graphcore ipu architecture via microbenchmarking},
  \href{http://arxiv.org/abs/1912.03413}{{\tt arXiv:1912.03413}}.

\end{thebibliography}\endgroup
\bibliographystyle{JHEP}

\clearpage

\appendix

\end{document}